\pdfoutput=1

\documentclass[11pt]{article}

\usepackage[final]{acl}

\usepackage{times}
\usepackage{latexsym}
\usepackage{url}
\usepackage{graphicx}
\usepackage{soul}
\usepackage{booktabs}
\usepackage{colortbl}
\usepackage{multirow}

\usepackage[T1]{fontenc}

\usepackage[utf8]{inputenc}

\usepackage{microtype}

\usepackage{inconsolata}
\usepackage{amsmath}

\usepackage{graphicx}
\usepackage{enumitem}
\definecolor{colorimagecommonskintone}{RGB}{129, 101, 79}
\definecolor{bwimagecommonskintone}{RGB}{176, 176, 176}

\title{Studying and Mitigating Biases in Sign Language Understanding Models}

\author{Katherine Atwell \\
  Northeastern University \\
  \texttt{atwell.ka@northeastern.edu} \\\And
  Danielle Bragg \\
  Microsoft Research \\
  \texttt{danielle.bragg@microsoft.com} \\\AND
  Malihe Alikhani \\
  Northeastern University \\
  \texttt{m.alikhani@northeastern.edu} \\}

\begin{document}
\maketitle

\begin{abstract}

Ensuring that the benefits of sign language technologies are distributed equitably among all community members is crucial. Thus, it is important to address potential biases and inequities that may arise from the design or use of these resources. Crowd-sourced sign language datasets, such as the ASL Citizen dataset, are great resources for improving accessibility and preserving linguistic diversity, but they must be used thoughtfully to avoid reinforcing existing biases.

In this work, we utilize the rich information about participant demographics and lexical features present in the ASL Citizen dataset to study and document the biases that may result from models trained on crowd-sourced sign datasets. Further, we apply several bias mitigation techniques during model training, and find that these techniques reduce performance disparities without decreasing accuracy. With the publication of this work, we release the demographic information about the participants in the ASL Citizen dataset to encourage future bias mitigation work in this space.
\end{abstract}
\section{Introduction}

\begin{figure}
    \centering
    \includegraphics[width=0.85\linewidth]{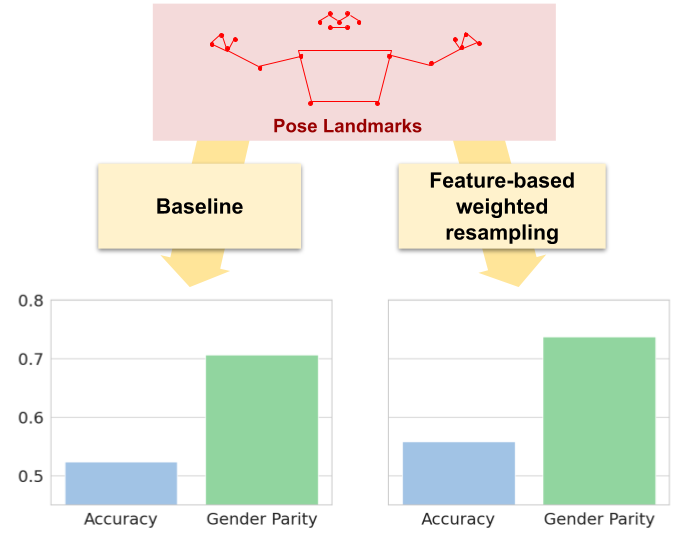}
    \caption{Accuracy and gender parity (calculated by dividing accuracy on female participants by accuracy on male participants) of the baseline pose-based ISLR model released with the ASL Citizen dataset (left) and our best-performing feature-based debiasing technique (right), in which we resample videos with lower video quality scores at a higher rate. Our approach improves both overall model accuracy and the gender parity.}
    \label{fig:intro-fig}
\end{figure}
Within the field of natural language processing, sign languages are under-resourced compared to spoken languages, compounded by the fact that most accessible information (e.g. online resources and social media) is written in a spoken language \cite{desai2024asl}. Datasets like the ASL Citizen dataset offer significant potential for improving accessibility and preserving the linguistic richness of sign languages, yet their use requires careful consideration to avoid reinforcing existing biases. In this context, our research aims to explore the factors that might influence the performance of models trained on these datasets, particularly when used for dictionary retrieval tasks.

Because sign languages have comparatively fewer resources than spoken languages, identifying biases in existing sign language resources is critical. But biases can manifest differently in sign languages than in spoken languages. For instance, ASL pronouns, unlike English pronouns, are not assigned a gender, so the common method of studying bias in English text through the lens of gendered pronoun use does not apply. Temporal elements, such as signing speed, also come into play, unlike in written language. Signing speed may be impacted by a signer's fluency, age, etc.

In this work, we analyze how signer demographics and more latent sources of bias may impact models trained on the ASL Citizen dataset for the task of Isolated Sign Language Recognition (ISLR). We first examine the demographic distributions in the ASL Citizen dataset, and present a linguistic analysis of the dataset based on the ASL-Lex \cite{caselli2017asl} annotations for each sign. We then report the prevalence of various linguistic and video-level features among demographics. We examine how demographic features, in conjunction with lexical and video-level features, may impact model results. Finally, we experiment with multiple debiasing techniques to reduce performance gaps between genders, and find that we are able to reduce these gaps and improve overall model accuracy (Figure \ref{fig:intro-fig}). 

In summary, we present an analysis of demographics, sign-level features, and video-level features in the ASL Citizen dataset and address the following research questions:
\begin{enumerate}[label=\textbf{\arabic*.}, nolistsep]
\item Which demographic and linguistic factors impact dictionary retrieval results for models trained on the ASL Citizen dataset?
\item Can we use debiasing strategies to mitigate disparate impacts while maintaining high performance for dictionary retrieval models?
\end{enumerate}

With this work, we also release the demographic data for the ASL Citizen dataset\footnote{Demographics available through the ASL Citizen project page: \url{https://www.microsoft.com/en-us/research/project/asl-citizen/}}, so future researchers can continue to study and mitigate bias in sign language processing systems. Further, we release the code for our experiments and analyses\footnote{\url{https://github.com/katherine-atwell/mitigating-biases-sign-understanding}}.
\section{Related Work}
Most readily-available information (i.e. online resources and social media) is written, which may limit accessibility for signers. Sign language processing tasks, such as dictionary retrieval, are designed to improve the accessibility of existing systems and resources for Deaf and Hard-of-Hearing (DHH) people. \citet{desai2024asl} created the ASL Citizen dataset to supplement existing dictionary retrieval resources with crowd-sourced videos from signers.

The ASL Citizen dataset was released to 1) address the resource gap between sign and spoken languages, and 2) improve \emph{video-based dictionary retrieval} for sign language, where signers demonstrate a particular sign and the system returns a list of similar signs, ranked from most to least similar. Video-based dictionary retrieval systems can help language learners understand the meaning of a sign, and allow signers to access dictionary resources using sign languages \cite{desai2024asl}. As a crowd-sourced dataset with videos of individual signs, the ASL Citizen dataset also serves to improve documentation of sign languages. This dataset is the first crowd-sourced dataset of videos for isolated signs, and members of deaf communities participated in, and were compensated for, this effort. When supplemented with the Sem-Lex benchmark \cite{10.1145/3597638.3608408}, a crowdsourced ISLR dataset released shortly after, 174k videos in total can be used for ISLR. The ASL Citizen dataset is licensed by Microsoft Research and is bound by the Microsoft Research Licensing Terms\footnote{Terms of use at \url{https://www.microsoft.com/en-us/research/project/asl-citizen/dataset-license/}. We are using this dataset in accordance with its intended use.}.

The ASL Citizen dataset is composed of videos of individual signs for isolated sign language recognition (ISLR). Other ISLR datasets with videos of individual signs have been released, including WL-ASL \cite{li2020word}, Purdue RVL-SLL \cite{wilbur2006purdue}, BOSTON-ASLLVD \cite{athitsos2008american}, and RWTH BOSTON-50 \cite{zahedi2005combination}. The above datasets, however, are not crowd-sourced. The closest dataset to the ASL Citizen dataset is the Sem-Lex Benchmark \cite{10.1145/3597638.3608408}, a crowdsourced ISLR dataset with over 91k videos. Because the Sem-Lex Benchmark does not release demographic information about the participants, we are not able to include it in our bias studies.


The ASL Citizen dataset is made up of crowd-sourced videos from ASL signers, where each video corresponds to a particular sign. The corpus is composed of videos for 2731 unique signs, all of which are contained in the ASL-Lex dataset \citet{caselli2017asl}, a lexical database of signs with annotations including the relative frequency, iconicity, grammatical class, English translations, and phonological properties of the sign. Thus, researchers studying this dataset can also take advantage of the ASL-Lex annotations. As part of the original data collection effort, demographic information about each participant was collected, but it was not released. With the publication of this work, we release the demographic data in this set, and provide a detailed analysis of this data. 

Our analyses of demographics and bias are motivated by evidence in the literature that a signer's demographics may impact their signing. For instance, characteristics of particular spoken languages or dialects have been shown to influence gestures, and in turn sign production \cite{cormier2010diversity}. One example of an ASL dialect is Black ASL, which scholarly evidence has shown to be its own dialect \cite{toliver2017investigating}, and for which documentation of dialectical differences dates back to 1965 \cite{stokoe1965dictionary}. Whether an individual speaks Black ASL is likely heavily influenced by their race or ethnicity. An example of geographic differences is Martha's Vineyard, an island off the coast of the United States, where an entire sign language emerged due to the high prevalence of DHH individuals in this community. Hearing and DHH people alike used this language to communicate until the mid-1900s \cite{kusters2010deaf}. There is also a distinct Canadian ASL dialect used by signers in English-speaking areas of Canada \cite{padden2010sign}, which is documented in a dictionary \cite{bailey2002canadian}. Age of language acquisition also impacts ASL production; delayed first-language acquisition affects syntactic knowledge for ASL signers \cite{boudreault2006grammatical} and late acquisition (compared to native acquisition) was found to impact sensitivity to verb agreement \cite{emmorey1995effects}.

Previous work also indicates the impact of certain visual and linguistic features on sign language modeling. Training an ISLR model to predict a sign and its phonological characteristics was found to improve model performance by almost 9\% \cite{kezar-etal-2023-improving}. \cite{10222729} find improved performance when using attention to focus on hand movements in sign videos. 

To our knowledge, there are no existing works that extensively study various sources of model bias on a crowd-sourced dataset of sign videos with labeled participant demographics. With this work, we aim to address this gap with a systematic analysis of the impact of various participant-level, sign-level, and video-level features, and experiment with debiasing techniques to reduce disparities in model performance.
\section{Data}
The ASL Citizen dataset is a crowd-sourced dataset containing 83,399 videos of individual signs in ASL from 52 different participants. The dataset contains 2731 unique signs that are included in the ASL-Lex \cite{caselli2017asl} dataset, a dataset with detailed lexical annotations for each sign. The authors of the original work report some demographic statistics, but the demographics of individual (de-identified) participants have not been released. Here, we provide a detailed report that includes demographic breakdowns and analyses of various linguistic and video features in the dataset, including the breakdown of these features by gender. We release the participant demographics with this work.

\subsection{Demographic Distributions}
\label{sec:demographics}
In total, the ASL Citizen dataset is comprised of 32 (61.5\%) women and 20 (38.5\%) men. 21 women are represented in the training set (60\%), 5 in the validation set (83\%), and 6 in the test set (55\%). The vast majority of participants report an ASL level of 6 or 7, as we show in Figure \ref{fig:asl-levels-and-regions} in Appendix \ref{sec:demographic-plots}. The participants also list their U.S. states. Using this information, we divide them into four regions based on the U.S. Census definitions \footnote{\url{ https://www2.census.gov/geo/pdfs/maps-data/maps/reference/us_regdiv.pdf}}: Northeast, Midwest, South, and West. More participants in the dataset are from the Northeast than any other region, as shown in Figure \ref{fig:asl-levels-and-regions} in Appendix \ref{sec:demographic-plots}. We also find that the age range of participants is skewed; participants in their 20s and 30s make up 32 of the 52 participants (see Figure \ref{fig:ages} in Appendix \ref{sec:demographic-plots}).

Participants did not note their ethnicity or race for this dataset. As such, to uncover potential biases related to the participants' perceived skin tone in their videos, we run the {\tt skin-tone-classifier} Python package from \citet{10.1111} on the frame with the first detected face in each video. We find that when we do not specify that the videos were in color, the classifier most often detects them as black and white. When we specify that the videos are in color, the most common skin tone detected (out of the default color palette used in \citet{10.1111}) is {\color{white}{{\sethlcolor{colorimagecommonskintone}\hl{\#81654f.}}}} Because the classifier most commonly detects as black and white, we also try specifying the video frames as being black and white. In this setting, the most common skin tone detected is {{\sethlcolor{bwimagecommonskintone}\hl{\#b0b0b0}}}, and the distribution differs from when the images are specified color images. 
We plot these results in Figure \ref{fig:skin-tones}.

\subsection{Sign and Video Features}
\label{sec:sign-and-video-features}

Because the ASL Citizen dataset is composed of signs from ASL-Lex \cite{caselli2017asl}, we can utilize ASL-Lex's lexical annotations for each sign. No works have studied these features in-depth on the ASL Citizen sign videos. We also analyze the video lengths, similarities and differences from the seed signer, and other video features.

\paragraph{Video Length}
We study the distribution of video lengths in order to better understand how video length may vary in this dataset. 
We find that the distribution of video lengths (s) is skewed left, with a longer tail on the right, as shown in Figure \ref{fig:video-lengths}. 

We also study whether video lengths vary, on average, for participants of different ages and genders. To account for differences between the signs depicted by participants (since participants did not all record the same signs), for each video, we calculate the number of standard deviations (SDs) the video length is away from the mean for all videos of that sign - in other words, we calculate the $z$-score at the sign level. We show this calculation in the equation below, where $v_{i}(s)$ represents the length of video $i$ depicting sign $s$.

\vspace{-1.5em}
\begin{align}
    z = \frac{v_{i}(s)-\mu_{s}}{\sigma_{s}}
\end{align}
\vspace{-1.5em}

We find that, while men on average record videos over .3 SDs longer than the mean, women on average record videos over .2 SDs shorter than the mean. Thus, compared to other videos with the same sign, women record shorter videos than men on average. We show these results in Figure 
\ref{fig:video-lengths-genders-ages}. Older participants, particularly those in their 70s, record longer videos on average (again, relative to other videos of the same sign) than younger participants. During manual inspection, we find older participants are more likely to have longer pauses before or after signing than younger participants, which may explain this gap. We also show these results in Figure \ref{fig:video-lengths-genders-ages}.

\paragraph{Sign Frequency}
The ASL Citizen dataset is comprised of 2731 signs from the ASL-Lex dataset \citet{caselli2017asl}, a dataset with expert annotations about properties of each sign including frequency of use, iconicity, and varying phonological properties. To collect sign frequency labels, deaf signers who use ASL were asked to rate signs from 1 to 7 in terms of how often they appear in everyday conversations, where 1 was ``very infrequently" and 7 was ``very frequently". 
We plot and compare the sign frequency distributions for the ASL Citizen dataset and the ASL-Lex dataset in Figure \ref{fig:sign-frequency}, and find that they are very similar.

We also find that there is little variation in average sign frequency for different genders. For male participants, the average sign frequency is 4.1592, while the average sign frequency for female participants is 4.1395, indicating that female participants chose slightly less frequently-occurring signs than men overall. 

\paragraph{Sign Iconicity}
The ASL-Lex dataset also contains crowd-sourced annotations for sign iconicity, where non-signing hearing annotators watch videos of a sign and evaluated how much they look like the sign's meaning from 1 (not iconic) to 7 (very iconic). We calculate an average iconicity of 3.378 in the ASL-Lex dataset, and 3.379 in the ASL Citizen dataset. We plot these distributions in Figure \ref{fig:sign-iconicity}, and again find that they are very similar.

We find average iconicity is 3.378 for women and 3.381 for men. This indicates that, as with frequency, there is only a slight difference, on average, between the iconicity of signs chosen by male and female participants. 
\section{Methods}
Here, we describe the baselines for our ISLR experiments, along with the experimental settings we use.

\subsection{Baselines}
For our experiments, we use the baseline I3D and ST-GCN models which were trained on the ASL Citizen dataset and released along with the dataset.\footnote{\url{https://github.com/microsoft/ASL-citizen-code}}. We describe the details of these models below.

\paragraph{I3D} 
The I3D model is a 3D convolutional network trained on the video frames themselves \cite{carreira2017quo}. As with the original ASL Citizen baselines, we train our I3D model on preprocessed video frames from the sign videos in the ASL Citizen training set. These videos are each standardized to 64 frames by skipping or padding frames depending on video length. Videos are then randomly flipped horizontally to imitate right- and left-handed signers. 

\subsection{ST-GCN}
The ST-GCN model is a temporal graph convolutional network trained on pose information \cite{yan2018spatial}. As with the original ASL Citizen baseline, we obtain pose representations for each frame using Mediapipe holistic \cite{lugaresi2019mediapipe}, with a set of 27 keypoints established by OpenHands \cite{selvaraj-etal-2022-openhands}. These keypoints are center scaled and normalized using the distance between the shoulder keypoints. The frames are capped at a maximum of 128, and random shearing and rotation transformations are applied during training for data augmentation.

\subsection{Experimental Settings}
All baselines are run on a Mac Studio with an Apple M2 Max chip and 64GB RAM.
\paragraph{I3D}
We use the same experimental settings as the I3D ASL Citizen baseline: 75 epochs maximum, learning rate of 1e-3, weight decay of 1e-8, an Adam optimizer and ReduceLRonPlateau scheduler with patience 5. As described in the ASL Citizen paper, we calculate the loss by averaging cross-entropy loss and per-frame loss.

\paragraph{ST-GCN} As with the original ASL-Citizen baseline, we train our ST-GCN model
for a maximum of 75 epochs using a learning rate of 1e-3, an Adam optimizer, and a Cosine
Annealing scheduler.

\begin{figure}
    \centering
    \includegraphics[width=0.35\textwidth]{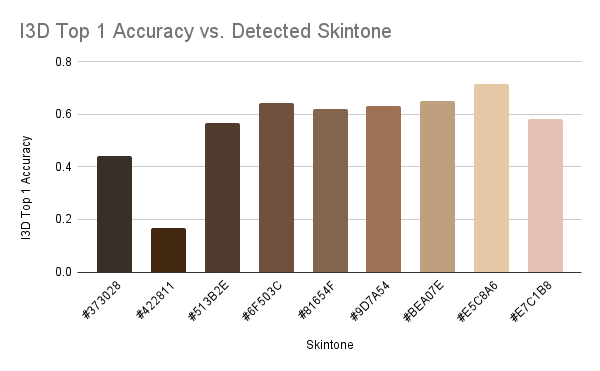} \\
    \includegraphics[width=0.35\textwidth]{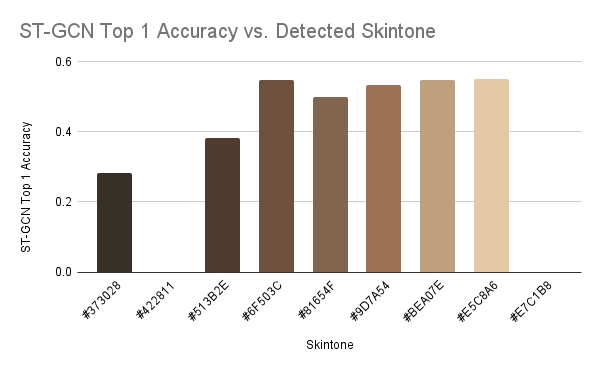}
    \caption{I3D (top) and ST-GCN (bottom) top 1 accuracy scores by detected skin tone. We find that, despite being less represented in the dataset, videos with lighter detected skin tones have higher accuracy scores on average for both models. The ST-GCN model, in particular, exhibits this behavior.}
    \label{fig:skintone-accuracy}
\end{figure}

\section{Which factors impact dictionary retrieval results in the ASL Citizen dataset?}
\subsection{Participant-level differences}
\label{sec:features-impacting-performance}

\paragraph{Baseline models perform over 10 percentage points better for male vs. female participants} We run the baseline I3D and ST-GCN models trained on the ASL Citizen dataset \cite{desai2024asl}, and, for both models, find an accuracy disparity between male and female participants. For the I3D model, the overall Top-1 accuracy is 0.6306, while for females it is 0.5914 and for males it is 0.6776; in other words, a gap of over 10 points in favor of male participants is observed. An even bigger gap is observed for the ST-GCN model; the overall Top-1 accuracy is 0.5944, while the Top-1 accuracy is 0.6838 for males and 0.52 for females. 

\paragraph{Average model accuracy varies greatly between participants}
One possible contributor to the above performance disparities for male and female participants is variation in participant-level model accuracy. There are 11 participants whose videos are in the test set for the ASL Citizen dataset. Of these 11 participants, 6 are female and 5 are male. When calculating accuracy scores for each participant, we find high variation for both models, with over 15-point differences between the highest and lowest accuracy scores (see Table \ref{tab:participant-accuracies}. This variation may contribute to the gender performance gap, as there are only a few participants of each gender in the test set. 

While performing manual inspection, we find several characteristics of user videos that appear to vary between participants. Different participants have different background or lighting quality, and some participants mouth the word being signed while other participants do not. We also find instances of repetition, where the sign is repeated in the video, from P15, a female participant. There are also some instances of fingerspelling, where participants fingerspell the sign before signing it. These and other individual differences may contribute to the observed performance disparities.

\paragraph{The models perform better on lighter skin tones than darker skin tones on average}

Despite darker skin tones making up most of the detected skin tones for videos in this dataset (see Figure \ref{fig:skin-tones}), we find that models average higher performance when the detected skin tone is lighter. We illustrate this phenomenon for both models in Figure \ref{fig:skintone-accuracy}. As this figure shows, I3D follows similar trends to ST-GCN in terms of comparative performance for different skin colors, performing the best for lighter skin tones \#BEA07E and \#E5C8A6. That being said, ST-GCN performs comparatively more poorly on the three darkest skin tones (\#373028, \#422811, and \#513B2E) and the lightest skin tone (\#E7C1B8) than I3D, when compared to the higher-performing skin tones. This indicates that, though both models show similar patterns regarding the skin tones with higher/lower performances, the RGB-based I3D model appears to perform better overall on darker skintones than the ST-GCN model. Although we find variations in accuracy between participants in the previous section, the skin tones are categorized at the video level. Thus, it is possible to see variation in predicted skin tone for different videos recorded by the same individual. The lighting quality of individual videos may be a confounder for these results. 

\paragraph{Trained models exhibit the highest average performance on participants in their 20s and 60s}
The ASL Citizen test set is made up of 11 individuals in their 20s, 30s, 50s, and 60s. We find that, as with gender, model accuracy varies for 
different age ranges; the highest accuracy scores were achieved for participants in their 20s and 60s. This could be influenced by the proportion of participants in their 20s in the training set. 

\subsection{Video-level differences}

\paragraph{Performance decreases as the video length diverges from the average}

For each sign video in the ASL Citizen dataset, we calculate the \emph{z}-score of its video length compared to other videos of the same sign. We then place these values into buckets: less than -2, -2 to -1, -1 to 0, 0 to 1, 1 to 2, and more than 2 SDs from the mean. We find that, on average, the videos farther away from the mean see decreased model performance compared to the videos closest to the mean. The results in full are in Table \ref{tab:video-length-accuracy}.

\begin{table}
    \centering
    \small
    \begin{tabular}{r|c|c}
    \toprule
 \textbf{Std. devs from mean} & \textbf{I3D Top-1} & \textbf{ST-GCN Top-1} \\
 \midrule
 \midrule
         $n<-2$ &  0.38462& 0.3846 \\
         \rowcolor[gray]{.9} $-2\leq n<-1$ &  0.5551& 0.4862 \\
         $-1\leq n<0$ &  0.648& 0.5888 \\
         \rowcolor[gray]{.9} $0\leq n<1$ &  \textbf{0.6704} & \textbf{0.6449} \\
         $1\leq n<2$ &  0.5727& 0.5878 \\
 \rowcolor[gray]{.9} $n>2$ & 0.3846&0.4668 \\
 \bottomrule
    \end{tabular}
    \caption{Top-1 accuracy scores for videos within a certain number of SDs away from the mean for videos of the same sign. For both models, videos with lengths closer to the mean yield better model performance.}
    \label{tab:video-length-accuracy}
\end{table}

\paragraph{Performance decreases when video quality degrades}

\begin{figure}
    \centering
    \includegraphics[width=0.25\textwidth]{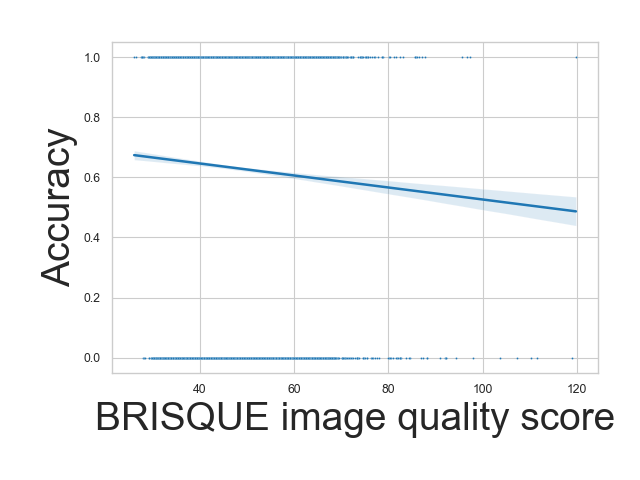}
    \caption{Association between BRISQUE image quality scores and accuracy. Higher BRISQUE scores indicate lower image quality, and vice versa. Thus, higher image quality appears to be associated with better model performance.}
    \label{fig:brisque-accuracy}
\end{figure}

In addition to video length, we study the impact of video quality on model accuracy. Given that we are studying the quality of individual video frames without a reference image, we use the BRISQUE score \cite{mittal2012no} to measure image quality of individual frames. Higher BRISQUE scores indicate lower quality, while lower BRISQUE scores indicate higher quality. We find that higher BRISQUE scores correlate negatively with Top-1 model performance for the I3D model, with a Spearman correlation of $\rho=-0.0367$ and a $p$-value of $p=1.53x10^{-8}$. We show a scatterplot of these results in Figure \ref{fig:brisque-accuracy}, along with a linear regression line.

\paragraph{Dissimilarity between participant and seed signer signs negatively impacts model accuracy for the ST-GCN pose model}

The Frech\'{e}t distance is often used as an evaluation metric for sign language generation, to study the similarity between generated signs and references \cite{hwang2024autoregressive, dong2024signavatar} (see \S~\ref{sec:frechet} for more details). In the ASL Citizen dataset, one of the participants is a paid ASL model who records videos for every sign, referred to as the ``seed signer". 

We study whether dissimilarity between the participant and seed signer may have a negative impact on model accuracy. To do so, we use the pose models used as input to the ST-GCN model. Every .25 seconds, we measure the distance between the model pose and the participant's pose at that frame, studying the distance between left hands and right hands separately. We find no significant relationship between right hand or left hand distance from the seed signer for the I3D model, and for the ST-GCN model we find a significant negative Spearman correlation between distance from the seed signer and accuracy for the right hand ($\rho=-.0289$, $p=0.001$). We plot these results, along with lines of best fit, in Figure \ref{fig:frechet-model-distances}.

\paragraph{When the average signing ``speed" is closer to the sign-level average, performance is better}

In addition to video length, we are interested in studying the average distance between poses over consistent time intervals. We want to study how much movement on average occurs within these increments, i.e. the ``speed" of sign production. We study this by calculating the pairwise Frechet distance between poses at each 0.25 second interval, with distance calculated between a pose and the pose .25s after, starting from the first frame. We again take this distance for the participants' right hand and left hand. We find that, on average, the farther away a participant's average signing speed is from that sign's mean, the worse performance is, with especially high performance degradations 2 SDs or more from the mean. We show these results in Table \ref{tab:speed-frechet-distance}.

\begin{table}
    \centering
    \small
    \begin{tabular}{r|cc|cc}
    \toprule
         &  \textbf{I3D} &  \textbf{ST-GCN} &  \textbf{I3D} & \textbf{ST-GCN} \\
         \textbf{SD from mean} & (LH) & (LH) & (RH) & (RH) \\
         \midrule
         \midrule
         $n<-2$ &  .4627&  .2139&  .5& .2375\\
         \rowcolor[gray]{.9} $-2\leq n<-1$ &  .6041&  .5804&  .6121& .5174\\
         $-1\leq n<0$ &  .6503&  .6426&  .6438& .6351\\
         \rowcolor[gray]{.9} $0\leq n<1$ &  .6244&  .5813&  .6423& .6145\\
         $1\leq n<2$ &  .6164&  .5261&  .616& .5744\\
         \rowcolor[gray]{.9} $n>2$ &  .5711&  0.4739&  .5619& .5107\\
         \bottomrule
    \end{tabular}
    \caption{Number of SDs away from the mean of the sign (in buckets) for the ``speed" of signing, i.e. the average Frechet distance between poses every 0.25 seconds, for right hand and left hand. We find that, for both right hand and left hand, the performance degrades as the average ``speed" of the sign production in a sign video deviates from the average for that particular sign.}
    \label{tab:speed-frechet-distance}
\end{table}

\subsection{Sign-level lexical features}
Here, we present results for four sign-level features annotated in the ASL-Lex dataset: sign frequency, iconicity, phonological complexity, and neighborhood density. We find that several of these features are significantly correlated with model performance, which we discuss below.

\paragraph{Sign frequency, phonological complexity, and neighborhood density are negatively correlated with model accuracy} As mentioned in \S~\ref{sec:sign-and-video-features}, sign frequency annotations in the ASL-Lex dataset were collected from ASL signers. The ASL-Lex 2.0 dataset \cite{sehyr2021asl} also contains a new phonological complexity metric. Using 7 different categories of complexity, scores were calculated by assigning a 0 or 1 to each category (depending on whether that category was present) and adding them together, for a maximum possible scores of 7 (most complex) and a minimum possible score of 0. The highest complexity score in the dataset was a 6. Neighborhood density was calculated based on the number of signs that shared all, or all but one, phonological features with the sign. 

Intuitively, we expect negative associations with phonological complexity and accuracy as well as neighborhood density and accuracy, and indeed find significant negative correlations ($\rho=-0.0618$, $p=0.005$ for phonological complexity and $rho=-0.0584$, $p=0.01$ for neighborhood density). However, we also find a significant negative association between sign frequency and model accuracy ($\rho=-0.057$, $p=0.011$). Existing work indicates that higher-frequency words are produced more quickly than low-frequency words \cite{jescheniak1994word, emmorey2013bimodal, Gimeno-Martinez2022Mar}; thus, it is possible that this association could be related to video length.  

\begin{figure*}
    \centering
    \includegraphics[width=.24\textwidth]{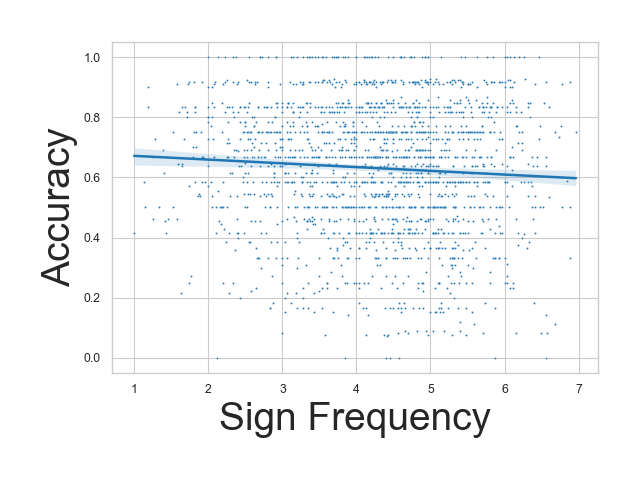}
    \includegraphics[width=.24\textwidth]{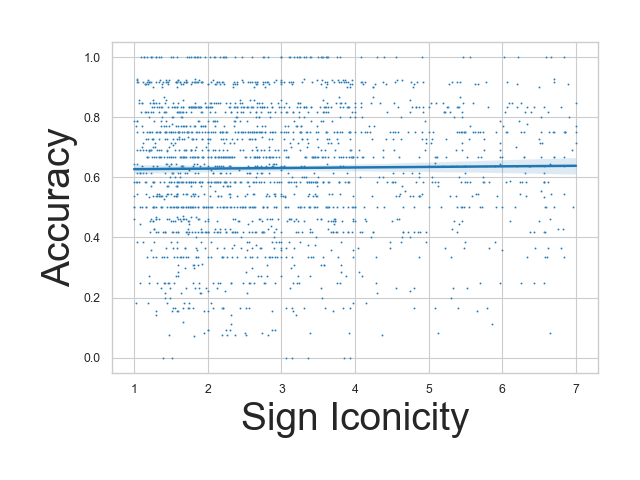}
    \includegraphics[width=.24\textwidth]{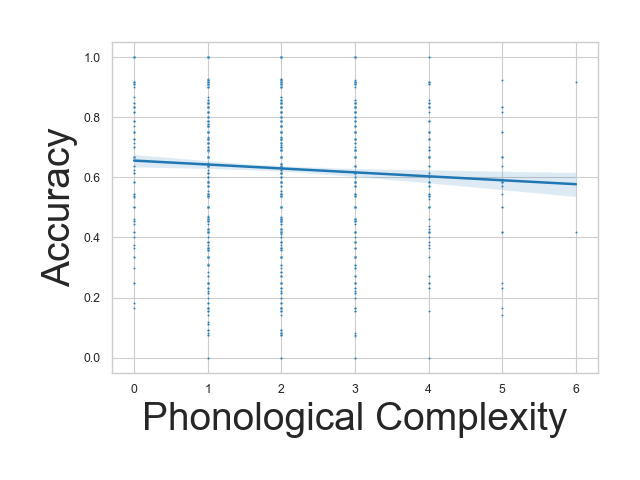}
    \includegraphics[width=.24\textwidth]{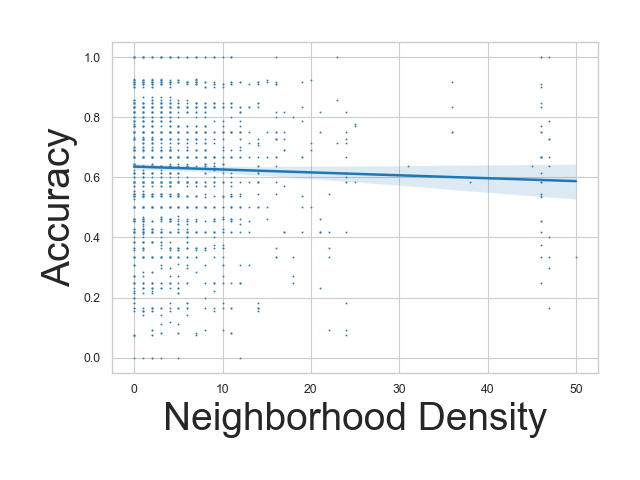}
    \caption{The relationships between sign frequency (left), sign iconicity (center left), phonological complexity (center right), and neighborhood density (right) and top 1 accuracy for the ST-GCN model. We find that sign frequency, phonological complexity, and neighborhood density are all significantly negatively correlated with model accuracy ($p<0.05$) when calculating the Spearman's rank correlation. However, despite a slight positive correlation between iconicity and accuracy, the $p$-value is not significant.}
    \label{fig:sign-features-accuracy}
\end{figure*}

\paragraph{There is no significant correlation between iconicity and model accuracy}
As mentioned in \S~\ref{sec:sign-and-video-features}, sign iconicity ratings were also collected for the ASL-Lex dataset. We find a very slight positive correlation between sign iconicity and model accuracy ($\rho=0.044$), which is not significant ($p=0.8424$). Thus, we conclude that visual similarity to the English word appears not to affect the model's ability to recognize a sign.

\subsection{Which features are the best predictors of model accuracy?}

After looking at the impacts of lexical, demographic, and video features on model accuracy, we are interested in studying which features are (by themselves) the best predictors of model accuracy. As such, we study the mutual information between each feature and the Top-1 accuracy for the I3D and ST-GCN models. We study 19 features in total, where some relate to participant demographics (e.g. age and gender), others relate to the sign lexical features (e.g. sign iconicity), and the rest are characteristics of individual videos (e.g. BRISQUE score and Frechet distances). We find that the 5 most impactful features are characteristics of individual videos (BRISQUE, Frechet from seed signer, and absolute $z$-score of ``signing speed"), with BRISQUE video quality scores showing the highest mutual information with Top-1 accuracy. Out of the lexical features, sign iconicity has the highest mutual information, and out of the demographic features, the participant's ASL level has the highest mutual information with the model performance. The results are in Table \ref{tab:mutual_information}.

\begin{table*}
    \centering
    \small
    \begin{tabular}{l|ccc|ccc|ccc|c}
    \toprule
 & \multicolumn{3}{c|}{\textbf{Overall}}& \multicolumn{3}{c|}{\textbf{Female participants}}& \multicolumn{3}{c|} {\textbf{Male participants}} & \textbf{Parity} \\
         Model&  Top-1&  Top-5&  Top-10& Top-1 & Top-5& Top-10& Top-1& Top-5&Top-10 & (Top-1) \\
         \midrule
         \midrule
         ST-GCN& .5238 & .7665 & .8295 & .4406 & .6886 & .7665 & .6236 & .8601 & \textbf{.9374} & .7065 \\
         ST-GCN (VL)& .5488 & .7923 & .8515 &.4666 & .7200 & .7941 & .6476 & .8791 & .9205 & .7205 \\
         ST-GCN (VL, fem.)& .5395 & .7926 & .8538 &  .4621& .7202 & .7974 & .63& \textbf{.8795} & .9216 & .7334 \\
         ST-GCN (brisque, HP)& .4723 & .7344 & .8046 & .3949 & .6551 & .7354 & 0.5653 & .8296 & .8877 & .6986\\
         ST-GCN (brisque, LP)& \textbf{.5580} & \textbf{.7960} & \textbf{.8545} & \textbf{.4801} & \textbf{.7279} & \textbf{.8011} & \textbf{.6516} & .8779 & .9187 & \textbf{.7368} \\
         \bottomrule
    \end{tabular}
    \caption{Performance of ST-GCN baseline against models that use the resampling strategies discussed in \ref{sec:resampling}. We find that all resampling strategies improve accuracy and gender parity over the baseline (for every metric but Top-10 Male), and resampling lower quality videos at a higher rate improves gender parity the most, followed closely by resampling based on video length from only female participants.}
    \label{tab:resampling}
\end{table*}

\section{Can we mitigate disparate impacts while maintaining high model performance for dictionary retrieval?}

\subsection{Training on single-gender subsets}
We first address the gender performance gap by training on participants of each gender in isolation. When doing this, we find a slight difference between the performance gaps for models trained on male-only and female-only subsets. For the model trained on the male-only subset, the Top-1 accuracy for male subjects is .292, and the Top-1 accuracy is .168. For the model trained on the female-only subset, the Top-1 accuracy for male subjects is .291, and the Top-1 accuracy for female subjects is .206. Thus, the model trained only on female subjects has a smaller gap, and higher accuracy parity, between male and female subjects than the model trained on only male subjects. However, both models have low performance overall, so the Top-1 accuracy parity for subjects of different genders (calculated by dividing the female accuracy by the male accuracy) is .7571 for the model trained on all subjects compared to .7079 for the model trained on only female subjects. The model trained on only male subjects has the lowest accuracy parity, at .5746. We show these results in Table \ref{tab:single-gender-subset-results} in Appendix \ref{sec:single-gender-subsets}.

\subsection{Training label shift}
In addition to training on single-gender subsets, we experiment with a label-shift approach to debiasing. Because ISLR is a multiclass problem, we experiment with the reduction-to-binary approach for debiasing multi-class classification tasks proposed by \citet{51666}.  We run the label-shift algorithm and train the ST-GCN model on the debiased labels for 25 epochs, and compare the performance of the debiased model to the ST-GCN model without debiasing, which we also train for 25 epochs. We find that the model trained on regular labels actually has a \emph{higher} ratio for female to male accuracy than the debiased model: .7476 for the baseline model, and .7052 for the debiased model. We show these results in full in Table \ref{tab:debias-label-shift-results}.

\subsection{Weighted resampling}
\label{sec:resampling}

Although there is a large gender performance gap observed (\S\ref{sec:features-impacting-performance}), based on the results from Table \ref{tab:mutual_information}, other features are much more heavily tied to model accuracy. Thus, it is likely that these features (in particular, features at the video level) may influence results. But what happens if the impact of videos with potentially-noisy features is reduced during training? We experiment with weighted resampling, where samples with certain features are more likely to be resampled. 
We explain how we calculate the resampling probability, and present results, for each variable we study in the paragraphs below.

\paragraph{Video length}
We first experiment with calculating the resampling probability based on video length. Given that videos closer to the mean produced higher accuracy scores, we wanted to resample these videos at a higher rate to reduce training noise. We calculate the probability of resampling as follows, where $v_i(s)$ refers to the length of video $i$ for sign $s$, $\mu_s$ refers to the mean video length of videos depicting sign $s$, and $\sigma_s$ refers to the SD for video lengths of videos depicting sign $s$:

\vspace{-1.5em}
\begin{align}
    P(resample) = \frac{1}{2^{\frac{v_i(s)-\mu_{s}}{\sigma_{s}}}}
\end{align}
\vspace{-0.7em}

We show the results for this approach in Table \ref{tab:resampling}, represented by the ST-GCN (VL) model. We find that this approach improves upon the baseline ST-GCN model by at least 2 percentage points for all accuracy metrics, and improves gender parity for Top-1 accuracy by 1.4\%. 

\paragraph{Video length for female participants}
We then experiment with the exact same resampling process described above, based on number of SDs from the mean for video length, but only resample videos from female participants. Because training on an all-female subset yielded a higher test accuracy for female subjects than an all-male subset (Table \ref{tab:single-gender-subset-results}), we want to investigate whether restricting our resampled data to female participants improves the gender performance gap. We show these results in Table \ref{tab:resampling}, under the baseline ST-GCN (VL, fem.). We find that this approach exceeds calculating the resampling probability using video length for participants of all genders for Top-5 and Top-10 accuracy. We also find that this baseline achieves the second-highest gender parity of all of the baselines, at 2.69\% higher than the baseline. Thus, we find evidence that resampling based on video length SDs, but only videos from female participants (the group with the lower model accuracy scores), greatly improves gender parity over the baseline model. 

\paragraph{BRISQUE score} Because the BRISQUE score shows the highest mutual information with Top-1 accuracy, we experiment with resampling based on the video quality. We experiment with two different resampling strategies: resampling higher-quality videos at a higher rate (\emph{resampling high quality}) and resampling lower-quality videos at a higher rate (\emph{resampling low quality}). We discuss these strategies below.

\noindent \textul{Resample high quality:} We first experiment with resampling more high-quality videos (lower BRISQUE scores) at a higher rate by setting the resampling probability as a function of the BRISQUE score, with higher BRISQUE scores reducing the resampling probability. We calculate the probability of resampling as follows, where $B_i(s)$ refers to the BRISQUE score of video $i$:

\vspace{-1.5em}
\begin{align}
    P(resample) = \frac{1}{2^{\frac{B_i}{100}}}
\end{align}
\vspace{-0.7em}

\noindent \textul{Resample low quality:}
We then experiment with resampling more low-quality videos (higher BRISQUE scores) at a higher rate by setting the resampling probability as a function that increases relative to the BRISQUE score. We calculate the probability of resampling as follows, where $B_i(s)$ refers to the BRISQUE score of video $i$:

\vspace{-1.5em}
\begin{align}
    P(resample) = \frac{1}{2^{\frac{100}{B_i}}}
\end{align}
\vspace{-0.7em}

Our results in Table \ref{tab:resampling} show that the latter approach, \emph{resample low quality}, achieves the highest overall accuracy and gender parity score.
\section{Conclusion}
In this work, we address a gap in sign language processing research by exploring biases in sign language resources, and experimenting with strategies to mitigate these biases. We focus on the ASL Citizen dataset in particular, and release demographic information for this dataset to aid future work. We find performance gaps related to skin tone, participant age, and gender. Still, we find that video level features, such as the video quality, signing ``speed", and video length, appear to be the best predictors of model accuracy. We find that selectively resampling data with video lengths closer to the mean improves overall performance. We also find that doing this resampling strategy for \emph{only} the group with lower model performance (female, when comparing genders) improves the gender parity for model performance. We find that resampling lower-quality videos at a higher rate achieves the highest Top-1 accuracy \emph{and} gender parity.
\section*{Limitations}

While in this work we find and document performance gaps between participants of different demographics such as age and gender, because of the differences between individual participants that we detail above (see Table \ref{tab:participant-accuracies}), and the number of participants in the test set (11), it is unclear how much of these differences are due to age or to other underlying factors.

Another limitation is that we focus on a single dataset. This is due in part to the fact that this is the only large-scale crowdsourced dataset for isolated sign language recognition with demographic labels. However, as more crowdsourced sign language resources become available, it is critical that these analyses are repeated on these datasets to assess the generalizability of our results.

\section*{Ethical Implications}

In our analysis of participant demographics, and accompanying features, for the ASL Citizen dataset, we present some characteristics of the dataset that vary between demographics. For instance, we discuss our findings that male participants and older participants typically record longer videos. It is important to emphasize that these findings should not be generalized to all ASL signers, and that they should instead be used to study the characteristics of this dataset in particular. 

Further, this work is not exhaustive; there are many sources of bias unexplored by this work, including differences in participant culture or ethnicity. There may be many more sources or dimensions of bias not covered in this paper that should be explored by future work. 

We also note that participants who chose to denote their demographic information (which was optional) consented for this information to be anonymously released as part of the dataset. No identifiable information about the participants will be released with the publication of this paper; rather, anonymous participant IDs will be accompanied with their demographics. 

\section*{Acknowledgments}

We would like to thank all of the participants who contributed videos to the ASL Citizen dataset, without whom this work would not have been possible.

\bibliography{anthology, acl_latex}

\appendix
\clearpage

\section{Participant Demographics}
\label{sec:demographic-plots}
Here, we plot the demographic information discussed in \ref{sec:demographics}. Note that providing demographic information was optional, so these numbers will not always add up to the total number of participants (52).

\begin{figure}
    \centering
    \includegraphics[width=.22\textwidth]{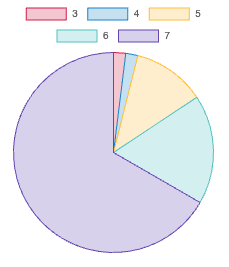}
    \includegraphics[width=.23\textwidth]{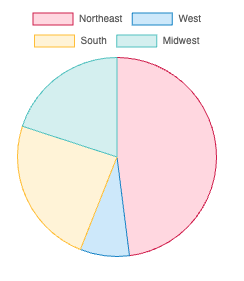}
    \caption{Distribution of ASL levels (left) and regions (right) of participants for the ASL Citizen dataset.}
    \label{fig:asl-levels-and-regions}
\end{figure}

In Figure \ref{fig:asl-levels-and-regions}, we plot the distribution of ASL levels and regions associated with the participants in the ASL Citizen dataset. We find that most participants are at an ASL level of 6 of 7, with only one participant each at level 3 or 4. A plurality of participants are from the Northeast, almost half. The West contains the fewest participants. 

\begin{figure}
    \centering
    \includegraphics[width=.4\textwidth]{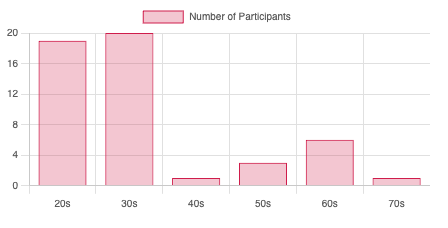}
    \caption{Age ranges of participants in the ASL Citizen dataset. Participants are skewed mostly towards their 20s and 30s, with a lesser skew towards participants in their 60s.}
    \label{fig:ages}
\end{figure}

In Figure \ref{fig:ages}, we plot the distribution of participants' ages. We find that participants are mostly skewed towards younger adults (20s and 30s) but that there is also a slight skew towards contestants in their 60s. Contestants in their 20s, 30s, 40s, 50s, 60s, and 70s are represented in the dataset, but contestants in their 40s and 70s are not represented in the test set.

\begin{figure}
    \centering
    \includegraphics[width=.4\textwidth]{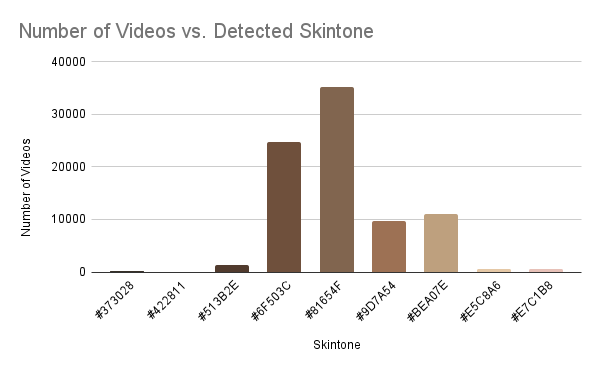} \\
     \includegraphics[width=.4\textwidth]{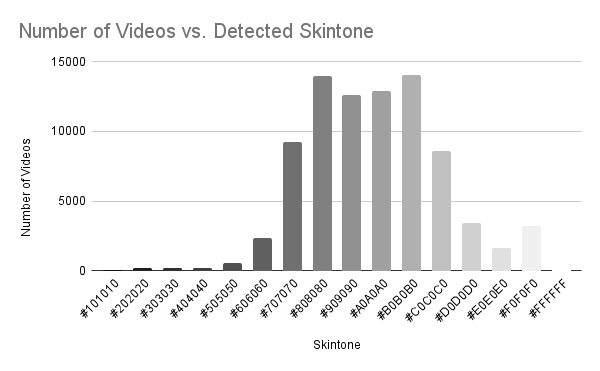}
    \caption{Frequency of detected skin tones of participants in videos when the video frames were set manually to color images (left) and black and white images (right)}
    \label{fig:skin-tones}
\end{figure}

In Figure \ref{fig:skin-tones}, we plot the distribution of skin tones in the dataset when frames are set as color images and black-and-white images. We include black-and-white images because we found that, when an image type was not set, the model detected the images as black-and-white images in the majority of cases. One notable finding is that the skin color model detected lighter skin tones more frequently when the images were set to black-and-white than when they were set to color images. This indicates possible unreliability of the skin color detection; it is possible, for instance, that when the images are set to color, the system classifies the skin colors as darker than they actually are. 

\section{Video Length Distributions}

\begin{figure}
    \centering
    \includegraphics[width=.48\textwidth]{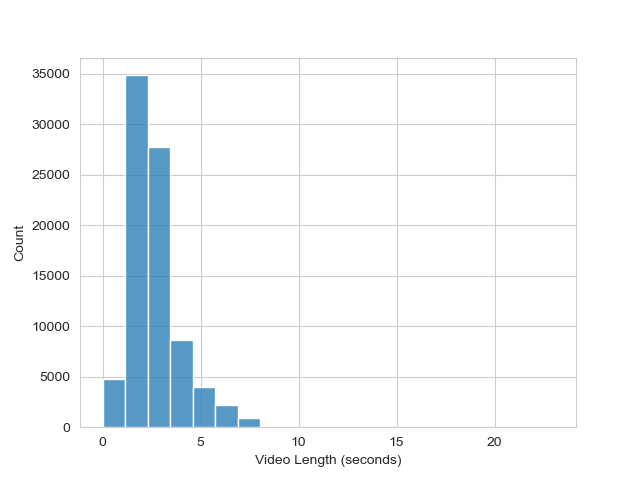}
    \caption{Distribution of video lengths for all sign videos in the ASL Citizen dataset. The distribution is skewed towards the right, with a long tail on the right.}
    \label{fig:video-lengths}
\end{figure}

In Figure \ref{fig:video-lengths}, we find that video lengths have a skewed distribution, where the average video length is higher than the median. In other words, video lengths lower than the mean are more common and vice versa, and there is a long tail to the right. After watching participants' videos, we suspect that this difference in video length is a result of some participants having a tendency to pause for multiple seconds at the beginning of end of their recording. This happens especially often with the first couple of videos that people record.

We also find that female participants have, on average, shorter videos related to their signs than male participants. For each sign video, we calculated the mean and standard deviation for all videos with that sign. We then calculated how many standard deviations those movies were away from the mean.

\begin{figure}
    \includegraphics[width=.48\textwidth]{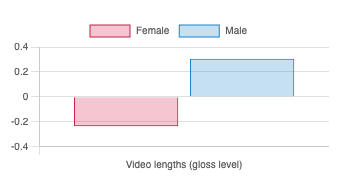} \\
    \includegraphics[width=.48\textwidth]{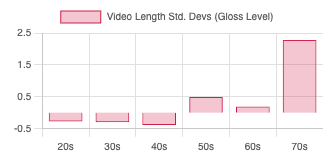}
    \caption{Average number of standard deviations away from the mean at the sign level for male and female participants (top) and participants in their 20s, 30s, 40s, 50s, 60s, and 70s (bottom). Relative to other videos of the same sign, women tend to record shorter videos, and older participants tend to record longer videos.}
    \label{fig:video-lengths-genders-ages}
\end{figure}
\section{Lexical Feature Distribution}

\begin{figure}
    \centering
    \includegraphics[width=.48\textwidth]{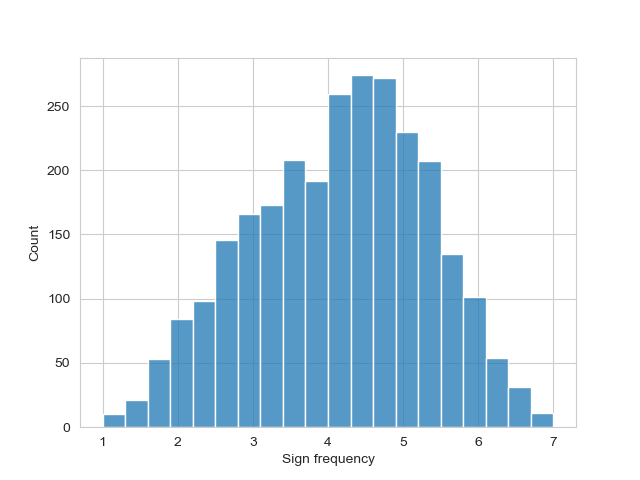}
    \includegraphics[width=.48\textwidth]{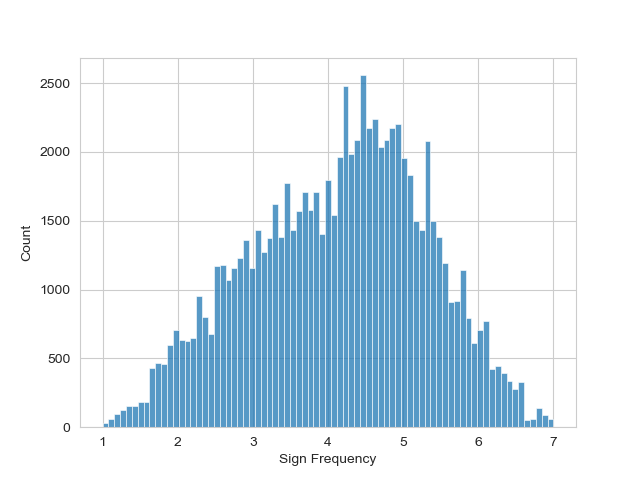}
    \caption{Distributions of labeled sign frequencies for each of the 2731 signs from the ASL-Lex dataset (top) and all of the sign videos in the ASL Citizen dataset (bottom). The distributions are very similar, indicating that users chosen signs of certain frequencies at a similar rate to how they are distributed in the ASL-Lex dataset.}
    \label{fig:sign-frequency}
\end{figure}

\begin{figure}
    \centering
    \includegraphics[width=.48\textwidth]{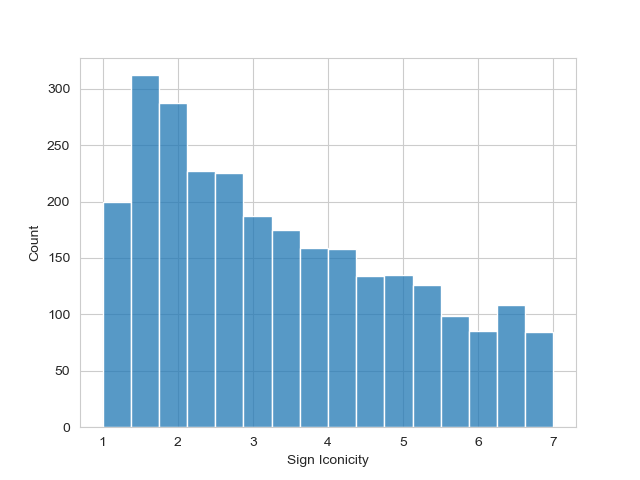}
    \includegraphics[width=.48\textwidth]{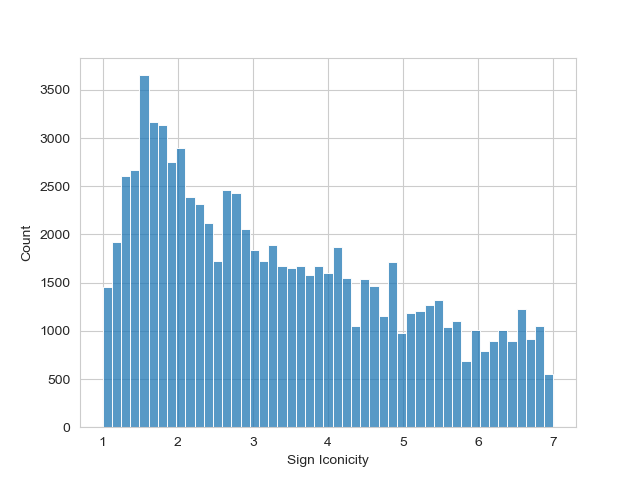}
    \caption{Distribution of sign iconicities in the ASL-Lex dataset (left) and the sign videos recorded in the ASL Citizen dataset (right). Like the sign frequencies, the iconicities in the ASL Citizen videos are distributed similarly to their distribution in the ASL-Lex dataset.}
    \label{fig:sign-iconicity}
\end{figure}

In addition to getting demographic and video features, we used the ASL-Lex \cite{caselli2017asl} annotations to analyze lexical features in the ASL Citizen dataset. We found that, for sign frequency and iconicity, the distributions are very similar to those in the ASL-Lex dataset. The distributions of both datasets are plotted side-by-side for frequency and iconicity, respectively, in Figures \ref{fig:sign-frequency} and \ref{fig:sign-iconicity}.

\section{Frech\'{e}t Distance}
\label{sec:frechet}

The Frech\'{e}t distance, used as a similarity metric between curves, and is commonly described in the following manner:

\begin{quote}
{\emph A man is walking a dog on a leash: the man can move on one curve, the dog on the other; both may vary their speed, but backtracking is not allowed. What is the length of the shortest leash that is sufficient for traversing both curves?} \\
- \cite{eiter1994computing}
\end{quote}
\section{Accuracies for different age ranges}
In Table \ref{tab:age_accuracies}, we show the Top-1 accuracy scores for the I3D and ST-GCN model for participants of different ages. We find the highest scores occur for participants in their 20s and 30s, with the third highest scores occuring for participants in their 60s. Participants in their 40s and 70s were not represented in the test set.

\begin{table}
    \centering
    \small
    \begin{tabular}{c|c|c|c}
    \toprule
         \textbf{Age range}&  \textbf{\# in test} &  \textbf{I3D Top-1} & \textbf{ST-GCN Top-1} \\
         \midrule
         \midrule
         20s&  2&  .6697& .6076\\
         \rowcolor[gray]{.9} 30s&  3&  .5689& .5336\\
         40s&  0 &  --& --\\
         \rowcolor[gray]{.9} 50s&  2&  .549& .5658\\
         60s&  3&  \textbf{.7016} & \textbf{.6421}\\
 \rowcolor[gray]{.9} 70s& 0& --&--\\
 \bottomrule
    \end{tabular}
    \caption{Average accuracy scores for participants of each age range in the test set. There were no participants in their 40s or 70s in the test set, and one participant did not specify their age. We find the highest performance in both models occurs for participants in their 20s and 60s.}
    \label{tab:age_accuracies}
\end{table}

\section{Model accuracies for each participant in the test set}

In Table \ref{tab:participant-accuracies}, we report the accuracy scores for the baseline ST-GCN model on the participants in the test set of the ASL Citizen dataset. We find differences of over 20 points between participant averages for both models. P6, P9, P15, P17, P18, and P22 disclosed that they are female, while the other participants disclosed that they are male.

\begin{table}
    \centering
    \small
    \begin{tabular}{c|c|c}
    \toprule
         Participant ID&  I3D Top-1&ST-GCN Top-1\\
         \midrule
         \midrule
         \textbf{P6} &  0.5456&0.4387\\
         \rowcolor[gray]{.9} \textbf{P9} &  0.6586&0.5663\\
         \textbf{P15} &  0.4653&0.5757\\
         \rowcolor[gray]{.9} \textbf{P17} &  0.6183&0.4997\\
         \textbf{P18} &  0.7065&0.5727\\
         \rowcolor[gray]{.9} \textbf{P22}&  0.5562&0.4671\\
         \textbf{P35} &  0.7204&0.7153\\
         \rowcolor[gray]{.9} \textbf{P42} &  0.6041&0.6949\\
         \textbf{P47} &  0.7471&0.7886\\
 \rowcolor[gray]{.9} \textbf{P48} & 0.6882&0.6652\\
 \textbf{P49} & 0.6327&0.556\\
 \bottomrule
    \end{tabular}
    \caption{Model top-1 accuracy scores on the set of videos recorded by each participant in the test set. For both models, there is high variation between participants, with scores ranging from 0.4653 to 0.7204 (I3D) and 0.4387 to 0.7886 (ST-GCN).}
    \label{tab:participant-accuracies}
\end{table}
\section{Frechet distance from seed signer}

\begin{figure}
    \centering
    \includegraphics[width=.23\textwidth]{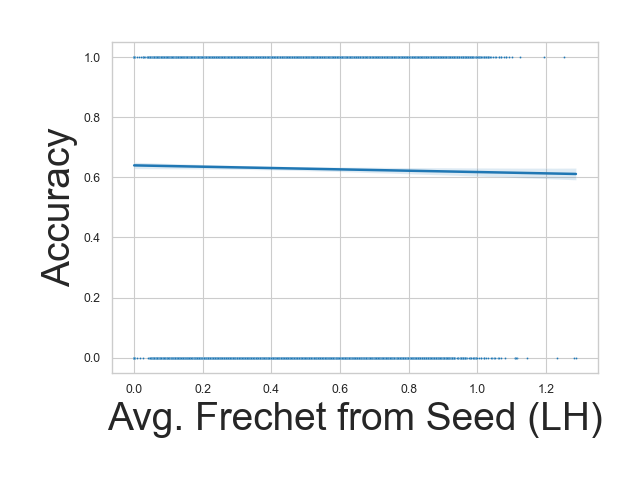}
    \includegraphics[width=.23\textwidth]{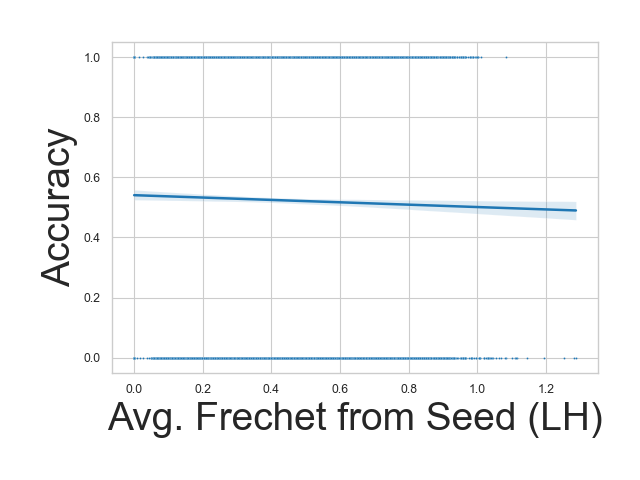}
    \includegraphics[width=.23\textwidth]{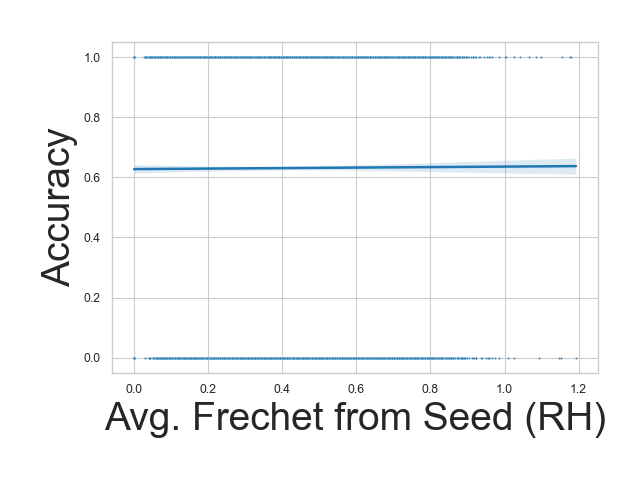}
    \includegraphics[width=.23\textwidth]{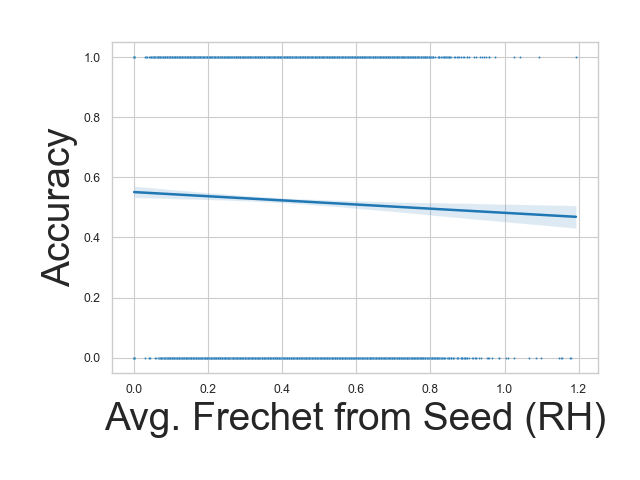}
    \caption{The Frechet distance from the seed (model) signer vs. top-1 accuracy for the I3D model (top) and ST-GCN model (bottom), with the distance between left hands on the left and the distance between right hands on the right.}
    \label{fig:frechet-model-distances}`    
\end{figure}

In Figure \ref{fig:frechet-model-distances}, we plot the Top-1 accuracies for the I3D and ST-GCN model as a function of the Frechet distance from the seed signer for each sign video (where the seed signer is a recruited ASL model for the ASL Citizen dataset). We find a significant negative correlation between Frechet distance from the seed signer and Top-1 accuracy for the ST-GCN pose model, but no significant correlations for the I3D model.

\section{Mutual Information Results}
\label{sec:mutual-information}

In Table \ref{tab:mutual_information}, we present the mutual information results in full for each studied variable. We study 19 variables total, spanning demographics, sign lexical features, and video-level features, and calculate the mutual information between each feature and the Top-1 accuracy. We find the highest levels of mutual information to occur for video-level features, suggesting features of individual videos are more impactful for model accuracy than demographic characteristics of the participants. Out of the demographic characteristics, the ASL level of the participant appears to be the most influential with respect to accuracy. 

\begin{table}
    \centering
    \small
    \begin{tabular}{l|c|c}
    \toprule
         \textbf{Feature} &  \textbf{Mut. Info} & \textbf{Mut. Info} \\
         & (ST-GCN) & (I3D) \\
         \midrule
         \midrule
         BRISQUE &  0.6920 & 0.6617 \\
         \rowcolor[gray]{.9} Avg. Frechet from seed (RH) &  0.6444 & 0.6217 \\
         Abs. Avg. Frechet SD (RH) &  0.6390 & 0.6090 \\
         \rowcolor[gray]{.9} Abs. avg. Frechet SD (LH) &  0.6285 & 0.5641 \\
         Avg. Frechet from seed (RH) &  0.5889 & 0.5403 \\
         \rowcolor[gray]{.9} Sign Iconicity &  0.0757 & 0.0508 \\
         Sign Frequency &  0.0619 & 0.0440 \\
         \rowcolor[gray]{.9} Abs. avg. Video Length SD &  0.0293 & 0.0399 \\
 ASL Level & 0.0048 &0.0020 \\
 \rowcolor[gray]{.9} Region & 0.0034 &0.0002 \\
 Neighborhood Density & 0.0032 &0.0026 \\
 \rowcolor[gray]{.9} Number Of Morphemes & 0.0026 &0.0012 \\
 Phonological Complexity & 0.0013 &0.0006\\
 \rowcolor[gray]{.9} Lexical Class & 0.0007 &0.0008 \\
 Iconicity Type & 0.0002 &0.0002 \\
 \rowcolor[gray]{.9} Gender & 0&0.0034 \\
 Age & 0&0.01107 \\
 Bounding Box Area (RH) & 0&0\\
 \rowcolor[gray]{.9} Bounding Box Area (LH) & 0&0\\
 \bottomrule
    \end{tabular}
    \caption{Mutual information for each of the features above and the Top-1 accuracy for the ST-GCN and I3D models, respectively. For both models, the BRISQUE score, average Frechet distance from the model (right hand and left hand) and the absolute value of the number of SDs of the average Frechet distance between frames are the top three features, with the other features far behind. This seemingly indicates that video-level features are the biggest indicator of model accuracy.}
    \label{tab:mutual_information}
\end{table}
\section{Results for models trained on single-gender subsets}
\label{sec:single-gender-subsets}

Here, we report the model results for the ST-GCN model trained on single-gender subsets, comparing models trained on all-male and all-female subsets to the model trained on all of the training data. In Table \ref{tab:single-gender-subset-results}, we report the Top-1, Top-5, and Top-10 accuracy scores for each model.

\begin{table*}
    \centering
    \small
    \begin{tabular}{l|ccc|ccc|ccc}
    \toprule
 & \multicolumn{3}{c|}{\textbf{Trained on female subjects}}& \multicolumn{3}{c|}{\textbf{Trained on male subjects}}& \multicolumn{3}{c}{\textbf{Trained on all subjects}}\\
         &   Top-1& Top-5& Top-10& Top-1& Top-5& Top-10& Top-1& Top-5& Top-10\\
         \midrule
\midrule
         All& 
     .244& .479& .581& .224& .434& .527& .594& .828& .881 \\
 Male& .291& .548& .653& .292 & .538& .639& .684& .902& .939 \\
 Female& .206& .421& .521& .168& .347& .433& .520& .767& .833 \\
 \bottomrule
 \end{tabular}

    \caption{Performances for ST-GCN model trained on only male subjects, only female subjects, and all subjects, respectively. We find that the model trained on only female subjects has the lowest performance gap between male and female subjects in the test set, but the ratio of female accuracy to male accuracy is highest for the model trained on all subjects.}
    \label{tab:single-gender-subset-results}
\end{table*}
\section{Results for model trained on debiased labels}
\label{sec:debiased-labels}

We report the results for a model trained for 25 epochs on training labels that were debiased using the reduction-to-binary techniques proposed by \citet{51666}. We find that the model trained on regular labels actually had a higher accuracy parity score (ratio of female accuracy to male accuracy) than the model trained on debiased labels. We show the Top-1, Top-5, and Top-10 results for each model in Table \ref{tab:debias-label-shift-results}.

\begin{table*}
    \centering
    \small
    \begin{tabular}{l|ccc|ccc|}
    \toprule
 & \multicolumn{3}{c|}{\textbf{ST-GCN}}& \multicolumn{3}{c|}{\textbf{ST-GCN (debiased)}}\\
         &   Top-1& Top-5& Top-10& Top-1& Top-5& Top-10\\
         \midrule
\midrule
         All& 
     .5323& .7997& .8622& .4821& .7576& .8265\\
 Male& .6173& .8781& .9254& .5746& .8493& .9014\\
 Female& .4615& .7343& .8096& .4052& .6811& .7641\\
 \bottomrule
 \end{tabular}

    \caption{Performances for ST-GCN model trained on regular training labels (left) and debiased training labels (right). We find that the accuracy parity, calculated as the ratio of female to male accuracy, is higher for the model trained on regular training labels than the debiased model.}
    \label{tab:debias-label-shift-results}
\end{table*}

\end{document}